\documentclass[11pt]{article}
\usepackage{acl2014}
\usepackage{times}
\usepackage{amsmath}
\usepackage{url}
\usepackage{latexsym}
\usepackage{algorithm}
\usepackage{algorithmic}
\usepackage{varwidth}
\usepackage{booktabs}
\usepackage{graphics}
\usepackage{multirow}
\usepackage{amssymb}
\usepackage[pdftex]{graphicx}

\DeclareMathOperator*{\argmin}{arg\,min}
\begin{document}
\title{Graph-Based Semi-Supervised Conditional Random Fields For Spoken Language Understanding Using Unaligned Data}

\author{Mohammad Aliannejadi, Masoud Kiaeeha, Shahram Khadivi,\\ \textbf{Saeed Shiry Ghidary}}

\author{Mohammad Aliannejadi \\
  Amirkabir University\\ of Technology\\
(Tehran Polytechnic)\\
{\tt m.aliannejadi@aut.ac.ir}\\\And
  Masoud Kiaeeha \\
Sharif University of\\
Technology\\
{\tt kiaeeha@ce.sharif.edu}\\\And
Shahram Khadivi \& \\ \textbf{Saeed Shiry Ghidary}\\
  Amirkabir University \\ of Technology\\
(Tehran Polytechnic)\\
{\tt \{khadivi, shiry\}@aut.ac.ir}\\}

\maketitle

\begin{abstract}
We experiment graph-based Semi-Supervised Learning (SSL) of Conditional Random Fields (CRF) for the application of Spoken Language Understanding (SLU) on unaligned data. 
The aligned labels for examples are obtained using IBM Model. 
We adapt a baseline semi-supervised CRF by defining new feature set and
altering the label propagation algorithm. 
Our results demonstrate that our proposed approach significantly improves the performance of the supervised model by utilizing the knowledge gained from the graph.
\end{abstract}

\section{Introduction}
The aim of Spoken Language Understanding (SLU) is to interpret the intention of the user's utterance.
More specifically, a SLU system attempts to find a mapping from user's utterance in natural language, to the limited set of concepts that is structured and meaningful for the computer.
As an example, for the sample utterance:\\
\makebox[0.4\textwidth][c]{\emph{I want to return to Dallas on Thursday}}\\
It's corresponding output would be:\\
\texttt{GOAL : RETURN\\
TOLOC.CITY = Dallas \\
RETURN.DATE = Thursday.\\
}
SLU can be widely used in many real world applications; however, data processing costs may impede practicability of it.
Thus, attempting to train a SLU model using less training data is a key issue.

The first statistical SLU system was based on hidden Markov model and modeled using a finite state semantic tagger employed in AT\&T's CHRONUS system \cite{pieraccini1992}.
Their semantic representation was flat-concept; but, later \newcite{he2005} extended the representation to a hierarchical structure and modeled the problem using a push-down automaton.
There are other works which have dealt with SLU as a sequential labeling problem. 
\newcite{ray2007} and \newcite{wang2006} have fully annotated the data and trained the model in discriminative frameworks such as CRF. 
CRF captures many complex dependencies and models the sequential relations between the labels; therefore, it is a powerful framework for SLU.

The Semi-Supervised Learning (SSL) approach has drawn a raft of interest among the machine learning community basically because of its practical application \cite{cha2006semi}.
Manual tagging of data can take considerable effort and time; however, in the training phase of SSL, a large amount of unlabeled data along with a small amount of labeled data is provided.
This makes it more practicable and cost effective than providing a fully labeled set of training data; thus, SSL is more favorable.

Graph-based SSL, the most active area of research in SSL in the recent years, has shown to outperform other SSL methods \cite{cha2006semi}.
Graph-based SSL algorithms are generally run in two steps: graph construction and label propagation.
Graph construction is the most important step in graph-based SSL; and, the fundamental approach is to assign labeled and unlabeled examples to nodes of the graph.
Then, a similarity function is applied to compute similarity between pairs of nodes.
The computed similarities are then assigned as the weight of the edges connecting the nodes \cite{zhu2003semi}.
Label propagation operates on the constructed graph.
Based on the constraints or properties derived from the graph, labels are propagated from a few labeled nodes to the entire graph.
These constraints include smoothness \cite{zhu2003semi,subra2010,talukdar2008weakly,garrette2013learning}, and sparsity \cite{das2012graph,zeng2013graph}.

Labeling unaligned training data requires much less effort compared to aligned data \cite{he2005}.
Nevertheless, unaligned data cannot be used to train a CRF model directly since CRF requires \emph{fully-annotated} data.
On the other hand, robust parameter estimation of a CRF model requires a large set of training data which is unrealistic in many practical applications.
To overcome this problem, the work in this paper applies semi-supervised CRF on unlabeled data.
It is motivated by the hypothesis that data is aligned to labels in a monotone manner, and words appearing in similar contexts tend to have same labels.
Under these circumstances, we were able to reach 1.64\% improvement on the F-score over the supervised CRF and 1.38\% improvement on the F-score over the self trained CRF.

In the following section we describe the algorithm this work is based on and our proposed algorithm.
In Section \ref{sec:results} we evaluate our work and in the final section conclusions are drawn.

\section{Semi-supervised Spoken Language Understanding}\label{sec:sslcrf}

The input data is unaligned and represented as a semantic tree, which is described in \cite{he2005}.
The training sentences and their corresponding semantic trees can be aligned monotonically; hence, we chose IBM Model 5 \cite{khadivi2005automatic} to find the best alignment between the words and nodes of the semantic tree (labels).
Thus, we have circumvented the problem of unaligned data. 
More detailed explanation about this process can be found in our previous work \cite{moli}.
This data is then used to train the supervised and semi-supervised CRFs.


\subsection{Semi-supervised CRF}
The proposed semi-supervised learning algorithm is based on \cite{subra2010}. 
Here, we quickly review this algorithm (Algorithm \ref{alg:1}).

\begin{algorithm}
  \caption{Semi-Supervised Training of CRF}
  \label{alg:1}
    \begin{algorithmic}[1]
    
\STATE $\Lambda_{(n=0)} = \mbox{TrainCRF}(\mathcal{D}_l)$
\STATE
      $\mbox{G} = \mbox{BuildGraph}(\mathcal{D}_l \cup \mathcal{D}_u)$ \\
\STATE
      $\{\mbox{r}\} = \mbox{CalcEmpiricalDistribution}(\mathcal{D}_l)$\\

\WHILE{not converged}
\STATE
      $\{\mbox{m}\} = \mbox{CalcMarginals}(\mathcal{D}_u, \Lambda_{n})$ \\
\STATE
      \{q\} = AverageMarginals(m) \\
\STATE
      $\{\hat{\mbox{q}}\}$ = LabelPropagation(q, r) \\
\STATE
      $\mathcal{D}_u^{v}$ = ViterbiDecode(\{$\hat{\mbox{q}}$\}, $\Lambda_{n}$)
\STATE
      $\Lambda_{n+1} = \mbox{RetrainCRF}(\mathcal{D}_l \cup \mathcal{D}_u^v, \Lambda_n);$
\ENDWHILE\label{mainwhile}\\
\STATE Return final $\Lambda_n$
   \end{algorithmic}
\end{algorithm}

In the first step, the CRF model is trained on the labeled data ($\mathcal{D}_l$) according to \eqref{eq:crf}:
\begin{equation}
\label{eq:crf}
\Lambda^\ast = \argmin_{\Lambda \in \mathbb{R}^{K}} \Big[ -\sum_{i=1}^{l} \log \mathrm{p}(\mathbf{y_i} | \mathbf{x_i};\Lambda) + \gamma \|\Lambda\|^2 \Big],
\end{equation}
where $\Lambda^\ast$ is the optimal parameter set of the base CRF model and $\|\Lambda\|^2$ is the squared $\ell_2$-norm regularizer whose impact is adjusted by $\gamma$. 
At the first line, $\Lambda^\ast$ is assigned to $\Lambda_{(n=0)}$ i.e. the initial parameter set of the model.

In the next step, the k-NN similarity graph (G) is constructed (line 2), which will be discussed in more detail in Section \ref{simgraph}. 
In the third step, the empirical label distribution (r) on the labeled data is computed. 
The main loop of the algorithm is then started and the execution continues until the results converge.

Marginal probability of labels (m) are then computed on the unlabeled data ($\mathcal{D}_u$) using Forward-Backward algorithm with the parameters of the previous CRF model ($\Lambda_n$), and in the next step, all the marginal label probabilities of each trigram are averaged over its occurrences (line 5 and 6).

In label propagation (line 7), trigram marginals (q) are propagated through the similarity graph using an iterative algorithm.
Thus, they become smooth.
Empirical label distribution (r) serves as the priori label information for labeled data and trigram marginals (q) act as the seed labels.
More detailed discussion is found in Section \ref{propagation}.

Afterwards, having the results of label propagation ($\hat{\mbox{q}}$) and previous CRF model parameters, labels of the unlabeled data are estimated by combining the interpolated label marginals and the CRF transition potentials (line 8). For every word position $j$ for $i$ indexing over sentences, interpolated label marginals are calculated as follows:
\begin{eqnarray}
\label{eq:viterbi}
\hat{\mathrm{p}}(y_i^{(j)} = y | \mathbf{x}_i) = \alpha \mathrm{p}(y_i^{(j)} = y | \mathbf{x}_i ; \Lambda_n) \nonumber \\ + ~ (1 - \alpha)\hat{\mathrm{q}}_{T(i,j)} (y),
\end{eqnarray}
where $T(i,j)$  is a trigram centered at position $j$ of the $i$th sentence and $\alpha$ is the interpolation factor. 

In the final step, the previous CRF model parameters are regularized using the labels estimated for the unlabeled data in the previous step (line 9) as follows:
\begin{eqnarray}
\label{crfretrain}
\Lambda_{n+1} = \argmin_{\Lambda \in \mathbb{R}^{K}} \Big[ -\sum_{i=1}^{l} \log \mathrm{p}(\mathbf{y_i} | \mathbf{x_i};\Lambda_n)\nonumber \\ ~~~-\eta \sum_{i=l+1}^{u} \log \mathrm{p}(\mathbf{y_i} | \mathbf{x_i}; \Lambda_n) + \gamma \|\Lambda\|^2 \Big],
\end{eqnarray}
where $\eta$ is a trade-off parameter whose setting is discussed later in Section \ref{sec:results}.

\subsection{CRF Features}
By aligning the training data, many informative labels are saved which are omitted in other works \cite{wang2006,ray2007}. 
By saving these information, the first order label dependency helps the model to predict the labels more precisely.
Therefore the model manages to predict the labels using less lexical features and the feature window that was [-4,+2] in previous works is reduced to [0,+2].
Using smaller feature window improves the generalization of the model \cite{moli}.

\subsection{Similarity Graph}\label{simgraph}
In our work we have considered trigrams as the nodes of the graph and extracted features of each trigram $x_2 ~ x_3 ~ x_4$ according to the 5-word context $x_1 ~ x_2 ~ x_3 ~ x_4 ~ x_5$ it appears in.
These features are carefully selected so that nodes are correctly placed in neighborhood of the ones having similar labels.
Table \ref{tab:features} presents the feature set that we have applied to construct the similarity graph.

\begin{table}
\renewcommand{\arraystretch}{1.1}
\centering
  \begin{tabular}{|c|c|}
   \hline
  \bfseries Description & \bfseries Feature \\
\hline\hline
  Context & $x_1\ x_2\ x_3\ x_4\ x_5$ \\
\hline
  Left Context & $x_1\ x_2$ \\
\hline
  Right Context & $x_4\ x_5$ \\
\hline
  Center Word in trigram & $\_\ x_3\ \_$ \\
\hline
Center is Class & $IsClass(x_3)$ \\
\hline
 Center is Preposition & $IsPreposition(x_3)$ \\
\hline
  Left is Preposition & $IsPreposition(x_2)$ \\
 \hline \end{tabular}
\caption{Context Features used for constructing the similarity graph} \label{tab:features}
\end{table}

 $IsClass$ feature impacts the structure of the graph significantly. 
 In the pre-processing phase specific words are marked as classes according to the corpus' accompanying database.
As an example, city names such as Dallas and Baltimore are represented as \emph{city\_name} which is a class type. 
 Since these classes play an important role in calculating similarity of the nodes, $IsClass$ feature is used to determine if a given position in a context is a class type. 
 
 Furthermore, prepositions like \emph{from} and \emph{between} are also important, e.g. when two trigrams like "\emph{from Washington to}" and "\emph{between Dallas and}" are compared.
 The two trigrams are totally different while both of them begin with a preposition and are continued with a class.
 Therefore, $IsPreposition$ feature would be particularly suitable to increase the similarity score of these two trigrams.
In many cases, these features have a significant effect in assigning a better similarity score.

To define a similarity measure, we compute the Pointwise Mutual Information (PMI) between all occurrences of a trigram and each of the features.
The PMI measure transforms the independence assumption into a ratio \cite{lin1998automatic,razmara2013}.
Then, the similarity between two nodes is measured as the cosine distance between their PMI vectors.
We carefully examined the similarity graph on the training data and found out the head and tail trigrams of each sentence which contain \emph{dummy} words, make the graph sparse.
Hence, we have ignored those trigrams.



\subsection{Label Propagation}\label{propagation}
After statistical alignment, the training data gets noisy. 
Hence, use of traditional label propagation algorithms causes an error propagation over the whole graph and degrades the whole system performance. 
Thus, we make use of the Modified Adsorption (MAD) algorithm for label propagation.

MAD algorithm controls the label propagation more strictly. 
This is accomplished by limiting the amount of information that passes from a node to another \cite{talukdar2010}. 
Soft label vectors $\hat{Y_v}$ are found by solving the unconstrained optimization problem in (\ref{eq:mad}):
\begin{eqnarray}\label{eq:mad}
\min_{\hat{Y}} \quad \sum_{l \in C}\Big[ \mu_{1} (Y_{l}-\hat{Y_{l}})^\top S~ (Y_{l}-\hat{Y_{l}})~ \nonumber \\
 + ~\mu_{2}\hat{Y_{l}}^\top L' \hat{Y_{l}}~ + ~ \mu_{3} \big\|\hat{Y_{l}} - R_{l} \big\|^2 \Big],
\end{eqnarray}
where $\mu_i$ are hyper-parameters and $R_l$ is the empirical label distribution over labels i.e. the prior belief about the labeling of a node. 
The first term of the summation is related to label score injection from the initial score of the node and makes the output match the seed labels $Y_l$ \cite{razmara2013}. 
The second term is associated with label score acquisition from neighbor nodes i.e. smooths the labels according to the similarity graph. 
In the last term, the labels are regularized to match a priori label $R_l$ in order to avoid false labels for high degree unlabeled nodes.
A solution to the optimization problem in \eqref{eq:mad} can be found with an efficient iterative algorithm described in \cite{talukdar2009new}.

Many errors of the alignment model are corrected through label propagation using the MAD algorithm; whereas, those errors are propagated in traditional label propagation algorithms such as the one mentioned in \cite{subra2010}.

\subsection{System Overview}
We have implemented the Graph Construction in Java and the CRF is implemented by modifying the source code of CRFSuite \cite{oka2007}. 
We have also modified Junto toolkit \cite{talukdar2010} and used it for graph propagation. 
The whole source code of our system is available online\footnote{\url{https://github.com/maxxkia/g-ssl-crf}}.
The input utterances and their corresponding semantic trees are aligned using GIZA++ \cite{och2000}; and then used to train the base CRF model. 
The graph is constructed using the labeled and unlabeled data and the main loop of the algorithm continues until convergence. 
The final parameters of the CRF are retained for decoding in the test phase.

\section{Experimental Results}\label{sec:results}
In this section we evaluate our results on Air Travel Information Service (ATIS) data-set \cite{dahl1994} which consists of 4478 training, 500 development and 896 test utterances. The development set was chosen randomly.
To evaluate our work, we have compared our results with results from Supervised CRF and Self-trained CRF \cite{yarowsky1995}.

\begin{table}
\begin{tabular}{c|c|c|c|}
\cline{2-4}
& \multicolumn{3}{ c| }{\% of Labeled Data} \\
\cline{2-4}
& 10 & 20 & 30 \\
\cline{2-4}
\cline{1-4}
\multicolumn{1}{|c|}{Supervised CRF} & $86.07$ & $87.69$ & $88.64$ \\
 \cline{1-4}
\multicolumn{1}{|c|}{Self-trained CRF} & $86.34$ & $87.73$ & $88.64$ \\
 \cline{1-4}
\multicolumn{1}{|c|}{Semi-supervised CRF} & $87.72$ & $88.75$ & $89.12$ \\
 \cline{1-4}
\end{tabular}
\caption{Comparison of training results. Slot/Value F-score in \%.} \label{tab:results}
\end{table}

For our experiments we set hyper-parameters as follows: for graph propagation, $\mu_1 = 1, \mu_2 = 0.01$, $\mu_3 = 0.01$, for Viterbi decoding, $\alpha = 0.1$, for CRF-retraining, $\eta = 0.1$, $\gamma = 0.01$.
We have chosen these parameters along with graph features and graph-related parameters by evaluating the model on the development set.
We employed the L-BFGS algorithm to optimize CRF objective functions; which is designed to be fast and low-memory consumer for the high-dimensional optimization problems \cite{bertsekas1999}.

We have post-processed the sequence of labels to obtain the slots and their values. The slot-value pair is compared to the reference test set and the result is reported in F-score of slot classification.
Table \ref{tab:results} demonstrates results obtained from our semi-supervised CRF algorithm compared to the supervised CRF and self-trained CRF.
Experiments were carried out having 10\%, 20\% and 30\% of data being labeled.
For each of these tests, labeled set was selected randomly from the training set.
This procedure was done 10 times and the reported results are the average of the results thereof.
The Supervised CRF model is trained only on the labeled fraction of the data.
However, the Self-trained CRF and Semi-supervised CRF have access to the rest of the data as well, which are unlabeled.
Our Supervised CRF gained 91.02 F-score with 100\% of the data labeled which performs better compared to 89.32\% F-score of \newcite{ray2007} CRF model.

As shown in Table \ref{tab:results}, the proposed method performs better compared to supervised CRF and self-trained CRF. The most significant improvement occurs when only 10\% of training set is labeled; where we gain 1.65\% improvement on F-score compared to supervised CRF and 1.38\% compared to self-trained CRF.

\section{Conclusion}\label{sec:conc}
We presented a simple algorithm to train CRF in a semi-supervised manner using unaligned data for SLU.
By saving many informative labels in the alignment phase, the base model is trained using fewer features.
The parameters of the CRF model are estimated using much less labeled data by regularizing the model using a nearest-neighbor graph.
Results demonstrate that our proposed algorithm significantly improves the performance compared to supervised and self-trained CRF.




%

\bibliographystyle{acl}
\bibliography{IEEEabrv,paper}

\end{document}